\documentclass[conference]{IEEEtran}
\IEEEoverridecommandlockouts
\usepackage{cite}
\usepackage{amsmath,amssymb,amsfonts}
\usepackage{algorithmic}
\usepackage{graphicx}
\usepackage{textcomp}
\usepackage{xcolor}
\usepackage{booktabs}   
\usepackage{graphicx}   
\usepackage{url}        
\usepackage{multirow}   
\usepackage{array}      
\usepackage{listings}   

\def\BibTeX{{\rm B\kern-.05em{\sc i\kern-.025em b}\kern-.08em
    T\kern-.1667em\lower.7ex\hbox{E}\kern-.125emX}}
\begin{document}

\title{ICME 2026 Grand Challenge on Cross-Scenario Defect Detection and Fine-Grained Severity Grading for High-Precision Manufacturing}


\author{

\IEEEauthorblockN{
Wei Sun$^{*}$\thanks{This challenge is organized by East China Normal University, Shanghai Jiao Tong University, Tsinghua Shenzhen International Graduate School, and Universit\'{e} de Nantes. Organizers are denoted by $^{*}$; all other authors are challenge participants.} \thanks{Website: \protect\url{https://idagrandchallenge.github.io/idachallenge.github.io}.},
Weixia Zhang$^{*}$,
Linhan Cao$^{*}$,
Mingkai Lu$^{*}$,
Xiongkuo Min$^{*}$,
\\
Xiaoping Zhang$^{*}$,
Patrick Le Callet$^{*}$,
Guangtao Zhai$^{*}$,
\\
Hongxing Chen,
Wenqi Wu,
Zhenhao Hu,
Shanshan Lin,
Guanjie Huang,
Kai Xie,
Rui Xin,
\\
Zilong Zhao,
Runmin Cong,
Ningjing Li,
Siqi Ma,
Yi Jin Ong,
Tianfei Zhou,
Shunzhou Wang,
\\
Zhiyang Chen,
Hao Fang,
Chen Zhang,
Tze-Hsiang Tang,
Dikai Li,
Xianjin Wu,
\\
Avinash Kumar Sharma,
Zhaoyang Wang,
Haiyong Chen,
Binyi Su,
Atik Shahariar
}
}

\maketitle

\begin{abstract}

This paper presents the IEEE International Conference on Multimedia \& Expo (ICME) 2026 Grand Challenge on Cross-Scenario Defect Detection and Fine-Grained Severity Grading for High-Precision Manufacturing. The challenge is motivated by two key limitations of existing industrial defect inspection systems: (1) current deep learning–based methods often suffer significant performance degradation when deployed in unseen production scenarios, and (2) most benchmarks neglect severity-aware assessment, which is critical for risk control and yield optimization. To address these limitations, we design two complementary tracks: \textit{Track~1 (Cross-Scenario Defect Detection)} targets accurate defect detection, localization, and classification across diverse unseen production environments; \textit{Track~2 (Fine-Grained Severity Grading)} requires assigning each detected defect an industry-standard severity level (\textit{i.e.}, \textit{Acceptable}, \textit{Marginal NG}, \textit{NG}, and \textit{Gross NG}). We construct a large-scale industrial dataset of high-resolution microscopic images spanning seven representative defect categories, comprising over 3,800 images with pixel-level instance annotations for Track~1 and over 2,600 images with severity-grade labels for Track~2. The challenge attracted 86 registered participants with 130 submissions; during the final testing phase, 21 teams submitted results and 12 teams provided models with technical fact sheets. The resulting benchmark, together with the diverse and effective solutions contributed by participating teams, sets a new standard for industrial defect analysis research.

\end{abstract}


\section{Introduction}
\label{sec:intro}

Automated visual inspection is a cornerstone of quality control in modern electronics manufacturing. While deep learning-based methods such as Mask2Former~\cite{cheng2022masked} have achieved remarkable progress under controlled settings, two fundamental limitations hinder their large-scale adoption in real-world factories. First, models trained on one production scenario often suffer from catastrophic performance degradation when transferred to unseen domains with different imaging devices or illumination conditions. Second, most existing methods treat defect detection as a binary or multi-class task, overlooking that defects of the same category can vary substantially in severity. In practice, distinguishing between an acceptable cosmetic blemish and a critical structural defect is essential for risk control and yield optimization, yet severity-aware assessment has received limited attention.

To address these gaps, we introduce the \textbf{Industrial Defect Analysis (IDA) 2026 Grand Challenge}, held in conjunction with ICME 2026, comprising two complementary tracks. \textbf{Track~1 (Cross-Scenario Defect Detection)} requires participants to accurately detect, localize, and classify defects across diverse and previously unseen production environments while minimizing false alarms, posing key technical challenges in domain adaptation and few-shot generalization. \textbf{Track~2 (Fine-Grained Severity Grading)} requires participants to assign each detected defect an industry-standard severity level (\textit{i.e.}, \textit{Acceptable}, \textit{Marginal NG}, \textit{NG}, and \textit{Gross NG}), posing challenges in fine-grained classification and ordinal regression.

\begin{figure}[t]
\centering
\includegraphics[width=\linewidth]{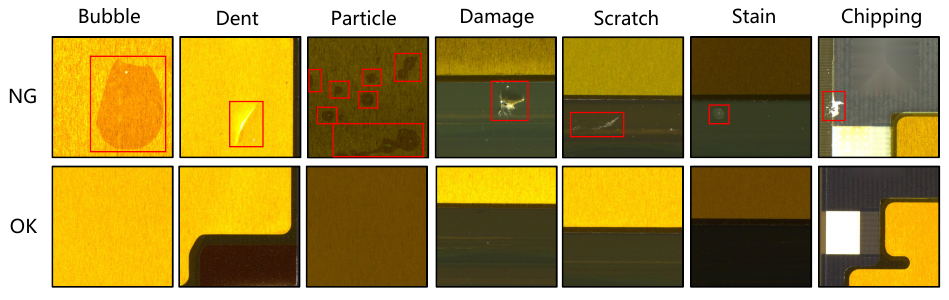}
\caption{Track~1 training dataset: representative defect samples across the seven categories with pixel-level instance annotations.}
\label{fig:track1_dataset_a}
\end{figure}

\begin{figure}[t]
\centering
\includegraphics[width=\linewidth]{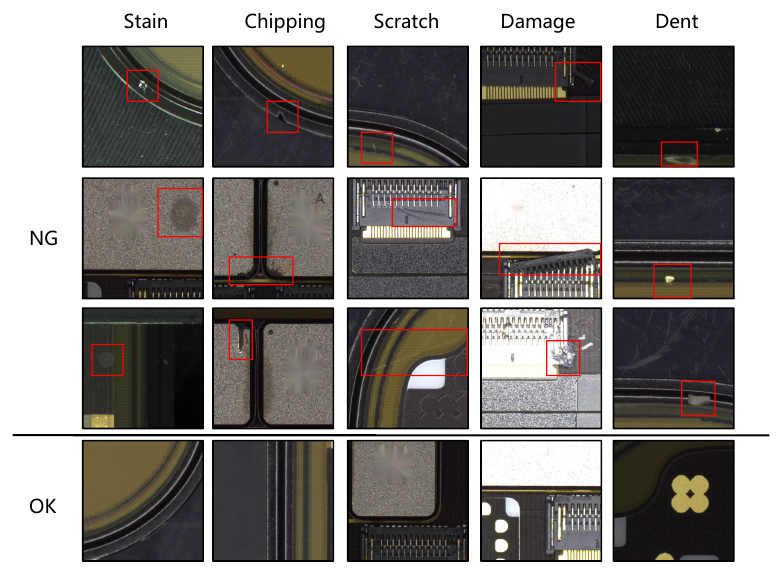}
\caption{Track~1 cross-scenario test set: examples illustrating domain shift through varied illumination, wafer texture, and defect scale.}
\label{fig:track1_dataset_b}
\end{figure}

\begin{figure}[t]
\centering
\includegraphics[width=\linewidth]{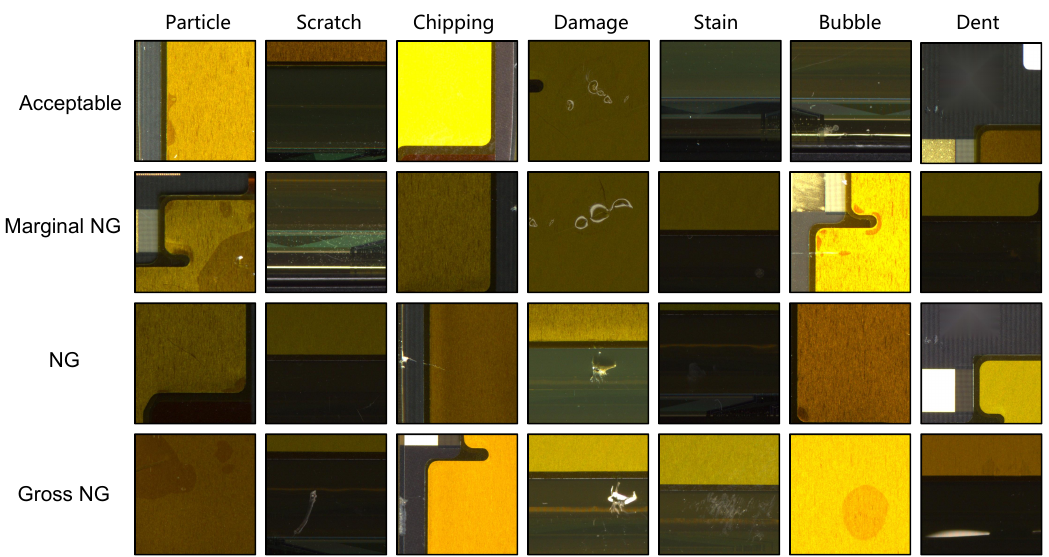}
\caption{Track~2 dataset: defect instances labeled with the four ordinal severity grades (\textit{Acceptable}, \textit{Marginal NG}, \textit{NG}, \textit{Gross NG}).}
\label{fig:track2_dataset}
\end{figure}

To support the challenge, we construct a large-scale industrial dataset of high-resolution microscopic images of semiconductor wafer surfaces, covering seven representative defect categories: \textit{Scratch}, \textit{Dent}, \textit{Particle}, \textit{Damage}, \textit{Stain}, \textit{Bubble}, and \textit{Chipping}. For Track~1, the training set contains 1,492 images (319 normal and 1,173 anomalous) with pixel-level instance annotations, while the test set comprises 2,323 images captured under more diverse and challenging conditions including variations in illumination, wafer texture, and defect scale. For Track~2, the training set shares the same images augmented with severity-grade labels, and the test set includes 1,156 anomalous images with severity annotations. The test sets are deliberately designed to introduce greater scene complexity, simulating real-world challenges of background texture interference, large intra-class morphological variation, and rare defect instances. Representative samples are illustrated in Fig.~\ref{fig:track1_dataset_a}, Fig.~\ref{fig:track1_dataset_b}, and Fig.~\ref{fig:track2_dataset}.

The challenge attracted 86 registered participants. During the development phase, 32 teams submitted 130 prediction results. In the final testing phase, 21 teams contributed submissions, and 12 teams provided their models along with technical fact sheets. In the following, we detail the challenge design, dataset, evaluation methodology, and provide an overview of the submitted methods and their performance.

\section{CHALLENGE DESIGN}

The IDA 2026 Grand Challenge aims to advance industrial defect analysis models that generalize across diverse production scenarios and perform severity-aware quality assessment. The challenge comprises two complementary tracks: Cross-Scenario Defect Detection and Fine-Grained Severity Grading.

\subsection{Task Definition}

\textbf{Track 1: Cross-Scenario Defect Detection.} Participants develop models to detect, localize, and classify surface defects on semiconductor wafer images from unseen production lines. The test set simulates real-world domain shift through variations in illumination, wafer texture, and defect scale absent from the training data.

\textbf{Track 2: Fine-Grained Severity Grading.} Beyond detection, participants assign each defect instance a severity level from four ordinal categories: \textit{Acceptable}, \textit{Marginal NG (Not Good)}, \textit{NG}, and \textit{Gross NG}, reflecting the risk a defect poses to product reliability.

\subsection{Dataset}

The dataset comprises high-resolution microscopic images of semiconductor wafer surfaces, covering seven defect categories: \textit{Scratch}, \textit{Dent}, \textit{Particle}, \textit{Damage}, \textit{Stain}, \textit{Bubble}, and \textit{Chipping}. Each image is annotated with instance segmentation masks, category labels, and severity grades where applicable.

For Track~1, the training set contains 1,492 images (319 normal, 1,173 anomalous) with 2,543 defect instances. The test set consists of 2,323 images (489 normal, 1,834 anomalous) with 1,921 instances, deliberately introducing greater scene diversity through varied lighting, texture patterns, and defect scales to evaluate cross-scenario generalization.

For Track~2, the training set shares the same images as Track~1 with additional severity labels. The test set contains 1,156 anomalous images with 1,388 defect instances, each labeled with one of the four severity levels.

\subsection{Evaluation Metrics}

For Track~1, the ranking is determined by:
\begin{equation}
S_{\text{Track1}} = 0.3 \times S_{\text{loc}} + 0.3 \times S_{\text{cls}} + 0.4 \times S_{\text{screen}},
\end{equation}
where $S_{\text{loc}}$ is the Mean Intersection over Union (mIoU), $S_{\text{cls}}$ is the Macro-F1 Score for defect classification, and $S_{\text{screen}}$ is a composite of image-level Recall and Specificity. The screening component receives the highest weight to reflect the industrial priority of minimizing false alarms.

For Track~2, the ranking is determined by:
\begin{equation}
S_{\text{Track2}} = 0.2 \times S_{\text{loc}} + 0.2 \times S_{\text{cls}} + 0.6 \times S_{\text{grade}},
\end{equation}
where $S_{\text{grade}}$ is measured by Quadratic Weighted Kappa (QWK), which penalizes disagreements proportionally to the squared distance between ordinal levels, aligning with practical risk implications.

\section{CHALLENGE RESULTS}

The challenge attracted broad participation from both academia and industry. After the final evaluation, 14 teams submitted valid results on Track 1 (Cross-Scenario Defect Detection) and 7 teams on Track 2 (Fine-Grained Severity Grading). Tables~\ref{tab:track1} and~\ref{tab:track2} summarize the final leaderboard for each track.

\begin{table}[t]
\scriptsize
\centering
\caption{Final leaderboard of Track 1: Cross-Scenario Defect Detection.}
\label{tab:track1}
\begin{tabular}{c l c c c c}
\toprule
Rank & Team & $S_{\text{loc}}$ & $S_{\text{cls}}$ & $S_{\text{screen}}$ & $S_{\text{Track1}}$ \\
\midrule
1 & FS-Seekers & 0.618 & 0.746 & 0.992 & 0.806 \\
2 & KZRR\_sdu & 0.628 & 0.665 & 1.000 & 0.788 \\
3 & trios & 0.467 & 0.590 & 1.000 & 0.717 \\
4 & MVP & 0.628 & 0.587 & 0.842 & 0.701 \\
5 & APC\_TDC & 0.561 & 0.638 & 0.827 & 0.690 \\
6 & accle & 0.628 & 0.555 & 0.555 & 0.577 \\
6 & LDK & 0.489 & 0.465 & 0.696 & 0.564 \\
6 & AIoTCTOCD & 0.724 & 0.399 & 0.441 & 0.513 \\
7 & zeyu & 0.693 & 0.417 & 0.442 & 0.510 \\
8 & jcjing & 0.532 & 0.460 & 0.524 & 0.507 \\
8 & EVA & 0.691 & 0.395 & 0.453 & 0.507 \\
9 & DefectsharpEye & 0.617 & 0.396 & 0.459 & 0.488 \\
10 & limzero & 0.772 & 0.247 & 0.255 & 0.408 \\
11 & MISSL & 0.335 & 0.308 & 0.500 & 0.393 \\
\bottomrule
\end{tabular}
\end{table}

\begin{table}[t]
\scriptsize
\centering
\caption{Final leaderboard of Track 2: Fine-Grained Severity Grading.}
\label{tab:track2}
\begin{tabular}{c l c c c c}
\toprule
Rank & Team & $S_{\text{loc}}$ & $S_{\text{cls}}$ & $S_{\text{grade}}$ & $S_{\text{Track2}}$ \\
\midrule
1 & MVP & 0.662 & 0.673 & 0.907 & 0.811 \\
2 & EVA & 0.689 & 0.667 & 0.885 & 0.802 \\
3 & Hebut AI & 0.652 & 0.658 & 0.891 & 0.797 \\
4 & accle & 0.684 & 0.675 & 0.873 & 0.796 \\
5 & jcjing & 0.684 & 0.674 & 0.872 & 0.795 \\
6 & limzero & 0.634 & 0.563 & 0.831 & 0.738 \\
7 & APC\_TDC & 0.546 & 0.666 & 0.790 & 0.716 \\
\bottomrule
\end{tabular}
\end{table}

\subsection{Track 1 Analysis}



Team FS-Seekers achieved the highest score of 0.806 with balanced performance across all sub-metrics. KZRR\_sdu and trios both attained perfect screening scores ($S_{\text{screen}}{=}1.000$), securing second and third place. A clear gap separates the top five from the rest, primarily driven by screening performance: the top five all exceeded 0.82 on $S_{\text{screen}}$, while lower-ranked teams generally fell below 0.56, confirming that false alarm control under domain shift is the primary bottleneck.

\subsection{Track 2 Analysis}


Team MVP achieved the top score of 0.811, driven by the highest severity grading quality ($S_{\text{grade}}{=}0.907$). The top five teams were tightly clustered within 0.016, indicating convergence on effective strategies. The primary differentiator was severity grading, where scores ranged from 0.790 to 0.907, confirming that ordinal boundary disambiguation remains challenging.

\section{PARTICIPATING METHODS: TRACK 1}
\label{sec:methods_t1}

\subsection{Team: FS-Seekers}
\textit{Members:} Hongxing Chen, Wenqi Wu, Zhenhao Hu, Shanshan Lin, Guanjie Huang (Shanghai Freesense Image Tech.).

\begin{figure}[t]
\centering
\includegraphics[width=\linewidth]{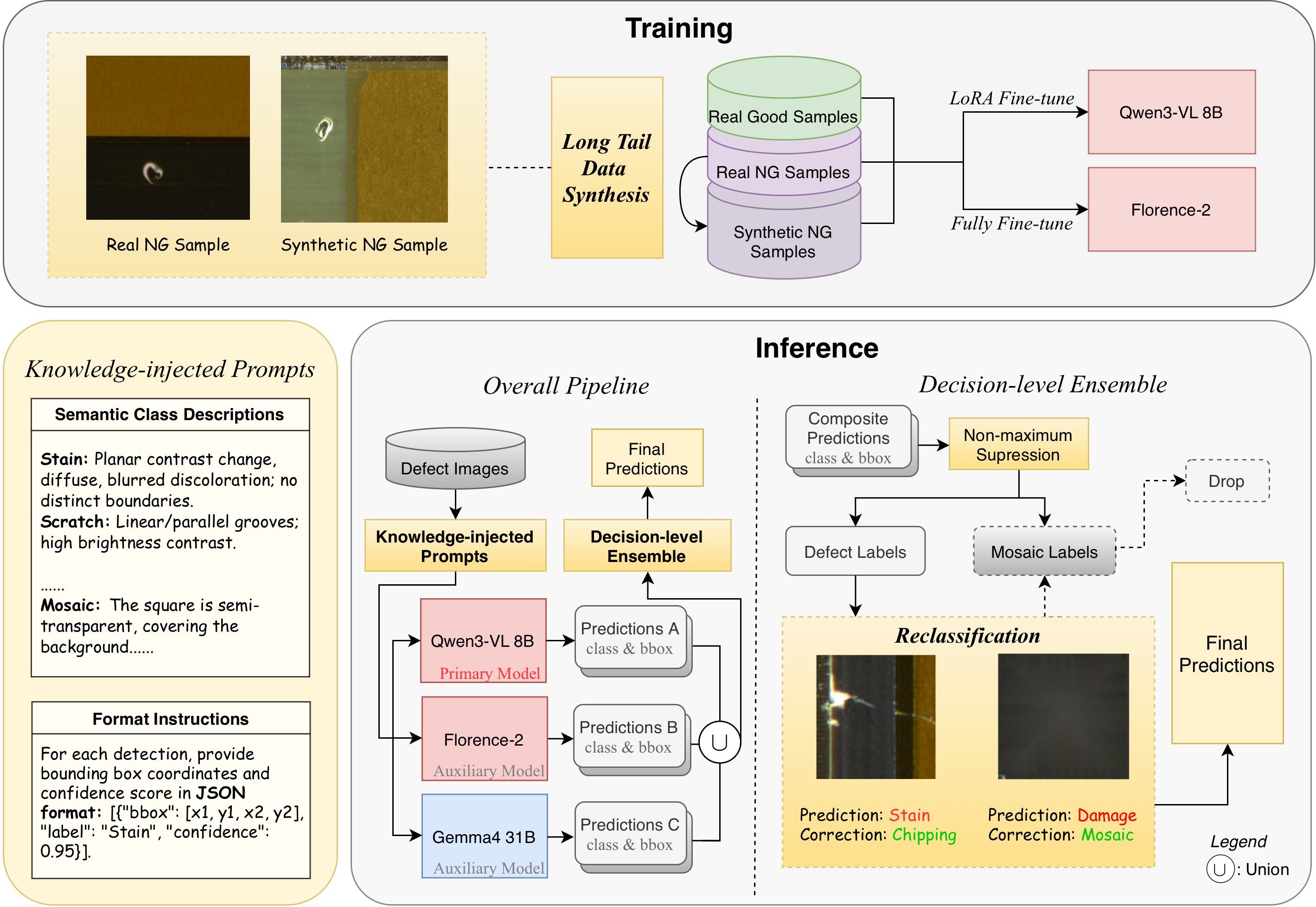}
\caption{The pipeline of the method proposed by Team FS-Seekers.}
\label{fig:fs_seekers}
\end{figure}

\textit{Method:} This team proposes a decision-level ensemble pipeline integrating three vision-language models (VLMs): LoRA-fine-tuned Qwen3-VL-8B~\cite{Qwen3VL} as the primary model, fully fine-tuned Florence-2 for fine-grained localization, and frozen Gemma-4-31B as an auxiliary semantic reasoner. All models share a unified knowledge-injected prompt that encodes industrial defect semantics such as morphological and textural descriptions. To mitigate severe class imbalance, DualAnoDiff~\cite{jin2024dualanodiff}, a category-conditional diffusion model, is employed to synthesize annotated samples for tail categories (202 for Dent and 185 for Damage). An inverse-square frequency weighting scheme is applied to the classification loss to amplify gradient signals for rare classes. During inference, predictions from all three models are aggregated via a union operation and refined through a three-stage cascaded post-processing pipeline: category-wise NMS, geometry-aware reclassification that corrects Chipping/Stain confusion by analyzing contour convexity and aspect ratio, and mosaic filtering based on Laplacian variance to discard artifacts. The overall pipeline is illustrated in Fig.~\ref{fig:fs_seekers}.

\subsection{Team: KZRR\_sdu}
\textit{Members:} Kai Xie, Rui Xin, Zilong Zhao, Runmin Cong (Shandong University).

\begin{figure}[t]
\centering
\includegraphics[width=\linewidth]{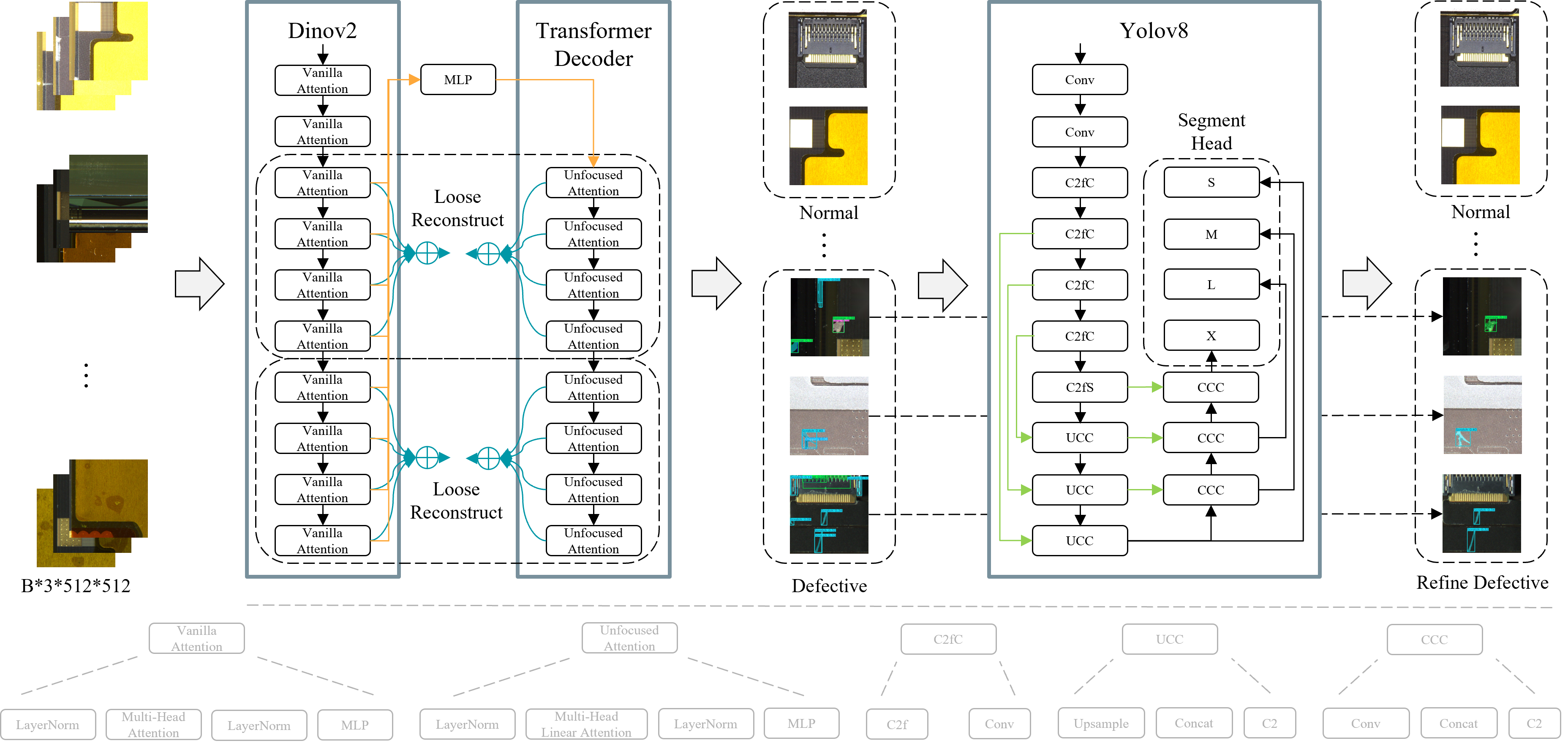}
\caption{The pipeline of the method proposed by Team KZRR\_sdu.}
\label{fig:kzrr}
\end{figure}

\textit{Method:} As illustrated in Fig.~\ref{fig:kzrr}, this team proposes a coarse-to-fine cascaded framework that decouples anomaly screening from fine-grained detection. The first stage employs Dinomaly~\cite{guo2025dinomaly} with a frozen DINOv2-Base~\cite{oquab2023dinov2} backbone to perform image-level anomaly scoring, where the score is computed as the mean response of the top 0.01\% highest-activation pixels. A symmetric padding strategy preserves original geometric structure at a fixed resolution of $518{\times}518$. Samples with anomaly scores below 0.05 are classified as defect-free, and candidate bounding boxes are extracted via connected-component analysis on the binarized anomaly map. The second stage forwards anomalous samples to a YOLOv8~\cite{yolov8} detector trained from scratch for 1,000 epochs at $512{\times}512$ for precise localization and classification. A dual-level fallback mechanism adopts YOLOv8 predictions when available and retains Dinomaly proposals otherwise, achieving a perfect screening score.

\subsection{Team: trios}
\textit{Members:} Ningjing Li, Siqi Ma, Yi Jin Ong, Tianfei Zhou (Beijing Institute of Technology); Shunzhou Wang (Henan University).

\begin{figure}[t]
\centering
\includegraphics[width=\linewidth]{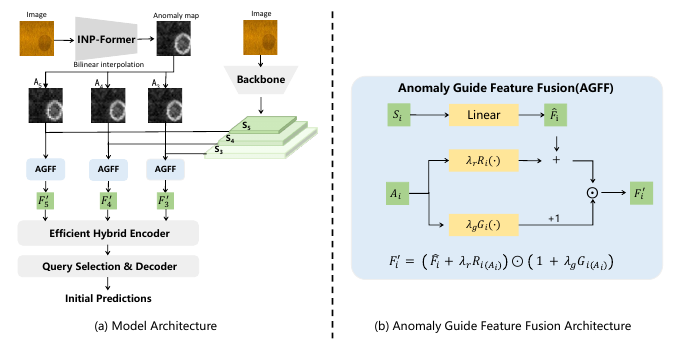}\\
{(a) Model architecture}\\[4pt]
\includegraphics[width=0.69\linewidth]{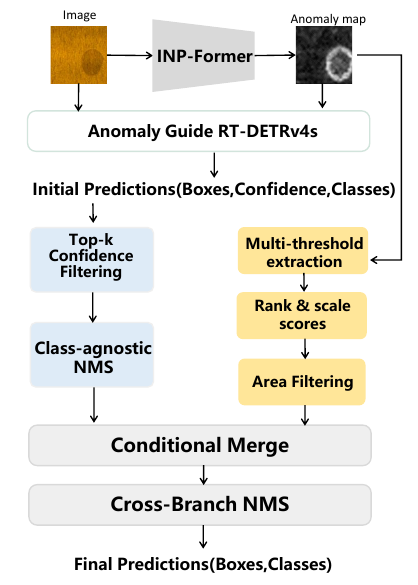}\\
{(b) Inference pipeline}
\caption{The pipeline of the method proposed by Team trios.}
\label{fig:trios}
\end{figure}

\textit{Method:} This team proposes an anomaly-guided detection framework that injects unsupervised anomaly priors directly into a supervised detector. The overall architecture is depicted in Fig.~\ref{fig:trios}(a). INP-Former~\cite{luo2025inpformer}, trained exclusively on defect-free images (720 of 901 normal samples), generates dense pixel-wise anomaly heatmaps by computing per-patch cosine similarity to learned normal prototypes from a frozen DINOv2~\cite{oquab2023dinov2} encoder. These heatmaps are integrated into the HybridEncoder of RT-DETRv4s~\cite{rtdetr} through a novel Anomaly-Guided Feature Fusion (AGFF) module with a dual-branch architecture: a residual branch for additive spatial bias and a gating branch for multiplicative modulation. Concretely, multi-scale feature maps $\{F_i\}$ extracted by an HGNetv2-S backbone are fused with bilinearly resized anomaly maps $\{A_i\}$ as $F'_i = (\hat{F}_i + \lambda_r R_i(A_i)) \odot (1 + \lambda_g G_i(A_i))$, with $\lambda_r{=}0.25$ and $\lambda_g{=}0.5$. The detector is initialized with COCO-pretrained weights and fine-tuned with Mosaic, MixUp, and multi-scale augmentation, while anomaly heatmaps remain unaugmented to maintain spatial alignment. The inference and post-processing pipeline (Fig.~\ref{fig:trios}(b)) operates as two parallel streams: the primary stream refines RT-DETRv4s predictions via top-$k$ confidence filtering and class-agnostic NMS, while an auxiliary stream generates anomaly-driven proposals through multi-threshold extraction, rank-and-scale ranking, and area filtering. The two branches are conditionally merged and consolidated by a final cross-branch NMS, recovering small or low-contrast defects that fall below standard confidence thresholds. The pipeline comprises only 10.4M parameters and 12.37G MACs, and completes training in approximately 1 hour on two A800 GPUs.

\subsection{Team: MVP}
\textit{Members:} Zhiyang Chen, Hao Fang, Chen Zhang (Shandong University).

\textit{Method:} This team adopts a direct fine-tuning approach based on the Mask2Former~\cite{cheng2022masked} instance segmentation framework with a Swin-Large~\cite{liu2021swin} backbone. The model is initialized from COCO-pretrained weights and fine-tuned for 2,500 iterations with large-scale jittering augmentation at $1024{\times}1024$ resolution on 2 A100 GPUs, completing training in approximately 30 minutes. No ensemble, external data, or dedicated anomaly screening module is employed. 

\subsection{Team: APC\_TDC}
\textit{Members:} Tze-Hsiang Tang (Schneider Electric Taiwan).

\textit{Method:} This team proposes a metric-aware multi-model ensemble combining four heterogeneous detectors: two CNN-style models (YOLO11l-seg and YOLO11x-seg~\cite{yolov8}) and two Transformer-style detectors (RF-DETR-Large and Co-DINO Swin-L). The architectural diversity is a deliberate design choice to yield complementary predictions. All models are pretrained on COCO and fine-tuned on the challenge data with varying input resolutions and augmentation strategies. Predictions are fused via Weighted Boxes Fusion (WBF)~\cite{solovyev2021weighted} with carefully tuned hyperparameters, including a dedicated score filter at 0.30 for Co-DINO to suppress low-confidence false positives. The team's key insight is that metric-aware confidence tuning, particularly optimizing the screening component, was the primary driver of performance gains. The total ensemble comprises approximately 436M parameters with training taking 6--7 hours.

\subsection{Team: LDK}
\textit{Members:} Dikai Li (Guangzhou University).

\textit{Method:} This team builds a single-model system on YOLO11 augmented with lightweight domain-generalization plug-in modules: IBN-a normalization~\cite{pan2018ibn} for style-invariant features, MixRand combining feature-statistic perturbation~\cite{li2022uncertainty} with progressive random convolutions, Space-to-Depth residual adapters for fine-grained spatial detail, and SimAM attention~\cite{yang2021simam}. The loss function replaces standard BCE with Varifocal Loss~\cite{zhang2021varifocalnet} and blends Normalized Wasserstein Distance~\cite{wang2021nwd} with CIoU for improved small-defect regression. A task-proxy-driven SWAD~\cite{cha2021swad} checkpoint selection strategy aligns weight averaging with the screening-oriented competition metric rather than generic mAP fitness. The model is trained for 2,000 epochs on a single RTX~4090 without ensemble or external data.

\subsection{Team: EVA}
\textit{Members:} Xianjin Wu (Huazhong University of Science and Technology).

\textit{Method:} This team frames the problem as a specificity-aware ensembling task, combining three ViT-Adapter~\cite{chen2022vitadapter} branches at different input resolutions (640\,px and 1024\,px) with a YOLO-World~\cite{cheng2024yoloworld} open-vocabulary branch. All branches are trained via a selection-then-refit protocol without external data, totaling approximately 287M parameters. Outputs are fused via WBF~\cite{solovyev2021weighted} and passed through a multi-stage decision pipeline including confidence thresholding, intra-class NMS, and class-specific prediction budgeting. The most distinctive component is a clean-image gate: if aggregated fused confidence falls below 0.15, the system outputs an empty defect list to suppress false alarms. The team found that test-time augmentation degraded screening behavior and was excluded from the final submission.

\subsection{Team: DefectSharpEye}
\textit{Members:} Avinash Kumar Sharma (Indian Institute of Technology Madras).

\textit{Method:} This team ensembles three independently trained detectors: RT-DETR-L~\cite{rtdetr} (32.0M), YOLOv8m (25.9M), and YOLOv8x~\cite{yolov8} (68.2M), combining Transformer and CNN architectures for complementary defect coverage. Domain-aware training strategies include 4--5$\times$ oversampling of rare defect classes and inclusion of 500 defect-free images with empty labels to teach normal appearance. Heavy augmentation is applied including rotation, scale jittering, mosaic, mixup, and copy-paste. During inference, predictions from all models are merged via NMS-based deduplication with a low confidence threshold of 0.10 to accommodate the confidence drop caused by domain shift.

\section{PARTICIPATING METHODS: TRACK 2}
\label{sec:methods_t2}

\subsection{Team: MVP}
\textit{Members:} Zhiyang Chen, Hao Fang, Chen Zhang, Runmin Cong (Shandong University).

\begin{figure}[t]
\centering
\includegraphics[width=\linewidth]{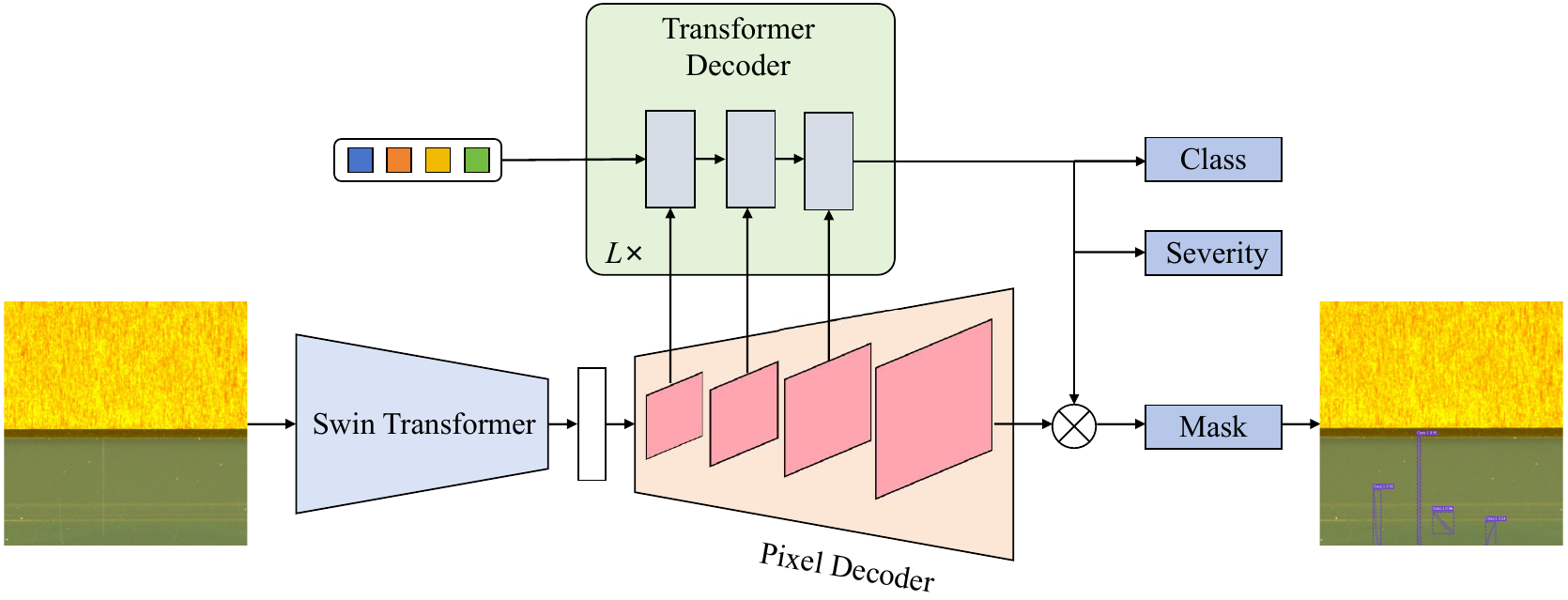}
\caption{The pipeline of the method proposed by Team MVP.}
\label{fig:mvp_t2}
\end{figure}

\textit{Method:} As shown in Fig.~\ref{fig:mvp_t2}, this team extends the Mask2Former~\cite{cheng2022masked} architecture with a parallel severity classification head added to the Transformer decoder, enabling end-to-end joint prediction of defect class, instance mask, and severity grade. For each decoder output, the model simultaneously predicts a class distribution, a severity distribution, and a binary mask, all trained through the same bipartite matching strategy. The severity head mirrors the design of the original class head, integrating the grading task without additional complexity. Training follows the same efficient protocol as their Track~1 submission: 2,100 iterations with a batch size of 8 on 2 A100 GPUs, completing in approximately 30 minutes. 

\subsection{Team: EVA}
\textit{Members:} Xianjin Wu (Huazhong University of Science and Technology).

\textit{Method:} This team adopts a decoupled two-stage design. The localization stage combines three architecturally diverse branches---YOLOv8x-seg~\cite{yolov8}, a DeiT-based transformer detector, and YOLO11x-seg---fused via WBF~\cite{solovyev2021weighted} with post-filter confidence thresholding, intra-class NMS, and a clean-image gate. Severity grading is performed only after localization has been consolidated: each fused defect instance is paired with the best overlapping raw mask, from which geometric and appearance descriptors are extracted and fed to a RandomForest classifier~\cite{breiman2001random}. This decoupling ensures that the grading module operates on a more stable geometric hypothesis than any single branch can provide, while remaining lightweight and easy to recalibrate.

\subsection{Team: Hebut AI}
\textit{Members:} Zhaoyang Wang, Haiyong Chen, Binyi Su, Atik Shahariar (Hebei University of Technology).

\begin{figure}[t]
\centering
\includegraphics[width=\linewidth]{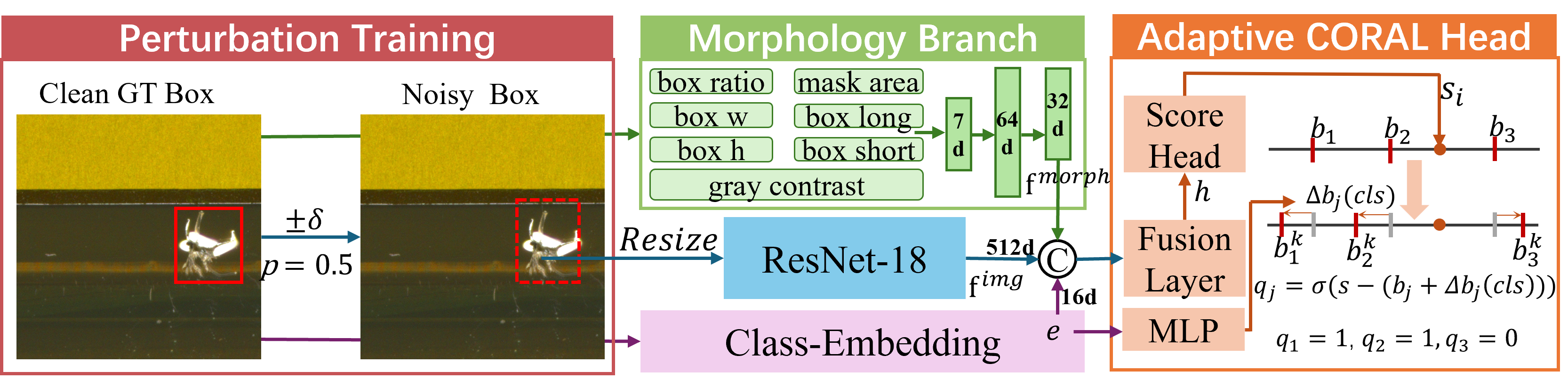}\\[26pt]
\caption{The pipeline of the method proposed by Team Hebut AI.}
\label{fig:hebutai}
\end{figure}

\textit{Method:} This team proposes Morphology-Aware Ordinal Learning (MAOL), a two-stage framework that decouples instance segmentation from severity grading (see Fig.~\ref{fig:hebutai}). A YOLOv8x-seg~\cite{yolov8} detector handles localization, after which each detected instance is processed by a ResNet-18~\cite{he2016deep} image encoder in parallel with seven handcrafted morphology descriptors (mask area, aspect ratio, box dimensions, and grayscale contrast) encoded by a lightweight MLP. The combined representation is passed to an adaptive CORAL~\cite{coral} ordinal head with class-conditional thresholds, which models severity through cumulative binary decisions to preserve the ordinal label structure. A prediction-aware perturbation strategy applied during training simulates realistic detector errors including translation, scale variation, and truncation, bridging the gap between clean training crops and noisy test-time predictions. The severity model contains only 11.47M parameters and trains in approximately 5 minutes.

\subsection{Team: APC\_TDC}
\textit{Members:} Tze-Hsiang Tang (Schneider Electric Taiwan).

\textit{Method:} This team extends their Track~1 four-model detection ensemble with a dedicated CORN~\cite{shi2021corn} ordinal regression head for severity prediction. For each fused defect instance, a $224{\times}224$ crop is extracted and encoded by an EfficientNet-B0 backbone, whose features are fed to the CORN head that models severity through conditionally ordered binary tasks, avoiding the rank-inconsistency problem of standard classifiers. The team's ablation study shows that replacing a rule-based severity baseline with the learned CORN head was the single largest contributor to performance, raising the local score from $S_2{=}0.441$ to $S_2{=}0.802$. A spatial sorting strategy based on defect center coordinates is applied to align outputs with the official evaluation pairing logic.

\section{Conclusion}

This paper presented a summary of the ICME 2026 Grand Challenge on Cross-Scenario Defect Detection and Fine-Grained Severity Grading for High-Precision Manufacturing, which aimed to benchmark industrial defect analysis methods capable of both cross-scenario generalization and fine-grained severity-aware quality assessment. The challenge attracted a range of solutions, including unsupervised anomaly pre-screening pipelines, vision-language model ensembles, end-to-end multi-task segmentation architectures, and decoupled ordinal regression frameworks. We hope that this challenge and its accompanying dataset will continue to support future research on industrial defect analysis, and serve as a useful benchmark for developing more robust, severity-aware, and deployable visual inspection systems for high-precision manufacturing.
\bibliographystyle{IEEEbib}
\bibliography{icme2026references}

\end{document}